\documentclass[sigconf]{acmart}
\usepackage{tabularx}
\usepackage{booktabs}
\usepackage{float}
\usepackage{microtype}
\usepackage{hyperref}
\usepackage{url}
\usepackage{amsmath}
\usepackage{graphicx}
\usepackage{longtable}
\usepackage{enumitem}
\usepackage{booktabs}
\usepackage{multirow}
\usepackage{tabularx, colortbl}
\usepackage{xcolor}
\usepackage{array, makecell}
\usepackage[normalem]{ulem}
\definecolor{correct}{HTML}{2ca02c}
\definecolor{wrong}{HTML}{d62728}
\definecolor{partial}{HTML}{ff7f0e}
\definecolor{noans}{HTML}{9467bd}
\definecolor{lightgray}{HTML}{F5F5F5}
\usepackage{amsmath}

\usepackage{amssymb}
\usepackage{float}

\AtBeginDocument{%
  }

\copyrightyear{2026}
\acmYear{2026}
\setcopyright{cc}
\setcctype{by}
\acmConference[CIKM '26]{Proceedings of the 35th ACM International Conference on Information and Knowledge Management}{November 07--11, 2026}{Rome, Italy}
\acmBooktitle{Proceedings of the 35th ACM International Conference on Information and Knowledge Management (CIKM '26), November 07--11, 2026, Rome, Italy}
\acmDOI{10.1145/3799682.3841078}
\acmISBN{979-8-4007-2539-5/2026/11}

\begin{document}

\title{GeoRisk-RAG: A Hierarchy-Aware Risk Framework for Improving RAG Reliability through Selective Answering}


\author{Meenu Ravi}
\affiliation{%
  \institution{Virginia Tech}
  \city{Alexandria}
  \state{Virginia}
  \country{U.S.}}
  \email{ravim@vt.edu}

\author{Shailik Sarkar}
\affiliation{%
  \institution{Florida Polytechnic University}
  \city{Lakeland}
  \state{Florida}
  \country{U.S.}}
  \email{ssarkar@floridapoly.edu}

\author{Lulwah AlKulaib}
\affiliation{%
 \institution{Kuwait University}
 \city{Kuwait City}
  \state{}
 \country{Kuwait}}
 \email{lalkulaib@cs.ku.edu.kw}

\author{Yordanos Tessema}
\affiliation{%
  \institution{Virginia Tech}
  \city{Alexandria}
  \state{Virginia}
  \country{U.S.}
  }
  \email{yordanost@vt.edu}

\author{Chang-Tien Lu}
\affiliation{%
  \institution{Virginia Tech}
  \city{Alexandria}
  \state{Virginia}
  \country{U.S.}}
  \email{ctlu@vt.edu}

\renewcommand{\shortauthors}{Ravi et al.}

\begin{abstract}
Current work on improving reliability in large language model (LLM)- generated answers has primarily leveraged Retrieval-Augmented Generation (RAG), knowledge-graph augmentation, and reinforcement learning. While these methods are adept at enhancing and measuring reliability through semantic similarity and faithfulness, they often struggle to distinguish semantic similarity from geographic validity. This is especially critical in natural hazard management domains where geographic granularity (i.e., town vs. city vs. state) is significant for decision-making, as responses valid in one municipality may not transfer to another. In such domains, a confidently wrong answer carries greater risk than abstaining. We present GeoRisk-RAG, a novel hierarchy-aware framework that addresses this geographic-validity gap through selective answering. This framework explicitly estimates geographic applicability using a Directed Acyclic Graph (DAG)-based distance for context retrieval before response generation. Experiments on a novel held-out wildfire-related question-answering (QA) dataset show that GeoRisk-RAG significantly reduces false confidence rates for location-dependent questions, lowering the rate to 0.009 compared with ~0.090 for standard semantic similarity and reranking baselines, while consistently achieving higher human preference alignment. This work provides a more comprehensive assessment of end-to-end RAG pipelines by integrating geographic validity and selective-answering behavior for safer decision-making in geospatial domains.
\end{abstract}

\ccsdesc[500]{Information systems~Retrieval effectiveness}


%
\keywords{RAG,
Location-Aware Retrieval,
Geospatial Information Retrieval,
Natural Hazard Management,
LLM Reliability}


\maketitle

\section{Introduction}
\begin{figure}
    \centering
    \includegraphics[width=0.8\linewidth]{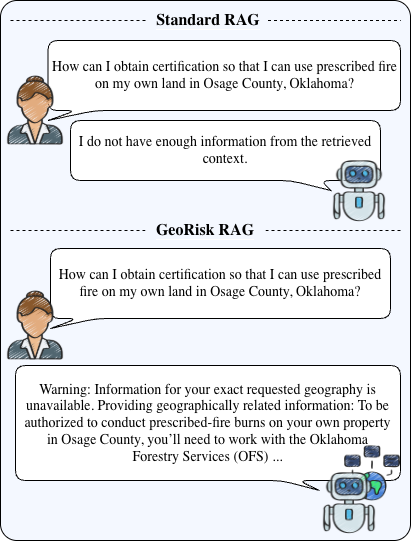}
    \caption{Example responses from Standard RAG (top) and GeoRisk-RAG (bottom) with the latter showing a warning+response despite a lack of geographical context for Osage County in the document-base, while the former abstains.}
    \label{fig:example}
\end{figure}
The increasing use of LLMs and other content generation tools has impacted users' interpretation and decision-making processes, particularly in domain-specific (e.g., geospatial or natural hazard) QA. Despite the plausible nature of their responses, LLMs face the continuing challenge of hallucinations and a lack of content reliability, compromising transparency, which is crucial for ensuring safety and trustworthiness.

To improve the groundedness and credibility of generated responses, RAG frameworks, often augmented with knowledge and real-world relationships expressed through graphs, have been increasingly adopted \cite{yuGraphRAGR1GraphRetrievalAugmented2026, lindersKnowledgeGraphextendedRetrieval2025, lewisRetrievalAugmentedGenerationKnowledgeIntensive2021}. Other approaches have focused on comparing the consistency of responses generated for a given query \cite{manakulSelfCheckGPTZeroResourceBlackBox2023} and the chain-of-thought prompting processes \cite{weiChainofThoughtPromptingElicits2023} across multiple runs. These dynamic methods of incorporating semantic relationships as contextual grounding and stochastic measurements for evaluating consistency have improved the reliability of LLM outputs in knowledge-retrieval tasks \cite{basuAdaptiveCheckGPTNovelSelfCorrecting2025}. In disaster-related QA, reliance on semantic or spatial proximity relevance alone is insufficient because evidence may be topically relevant yet geographically invalid for the target region.
\begin{figure*}
    \centering
    \includegraphics[width=1\linewidth]{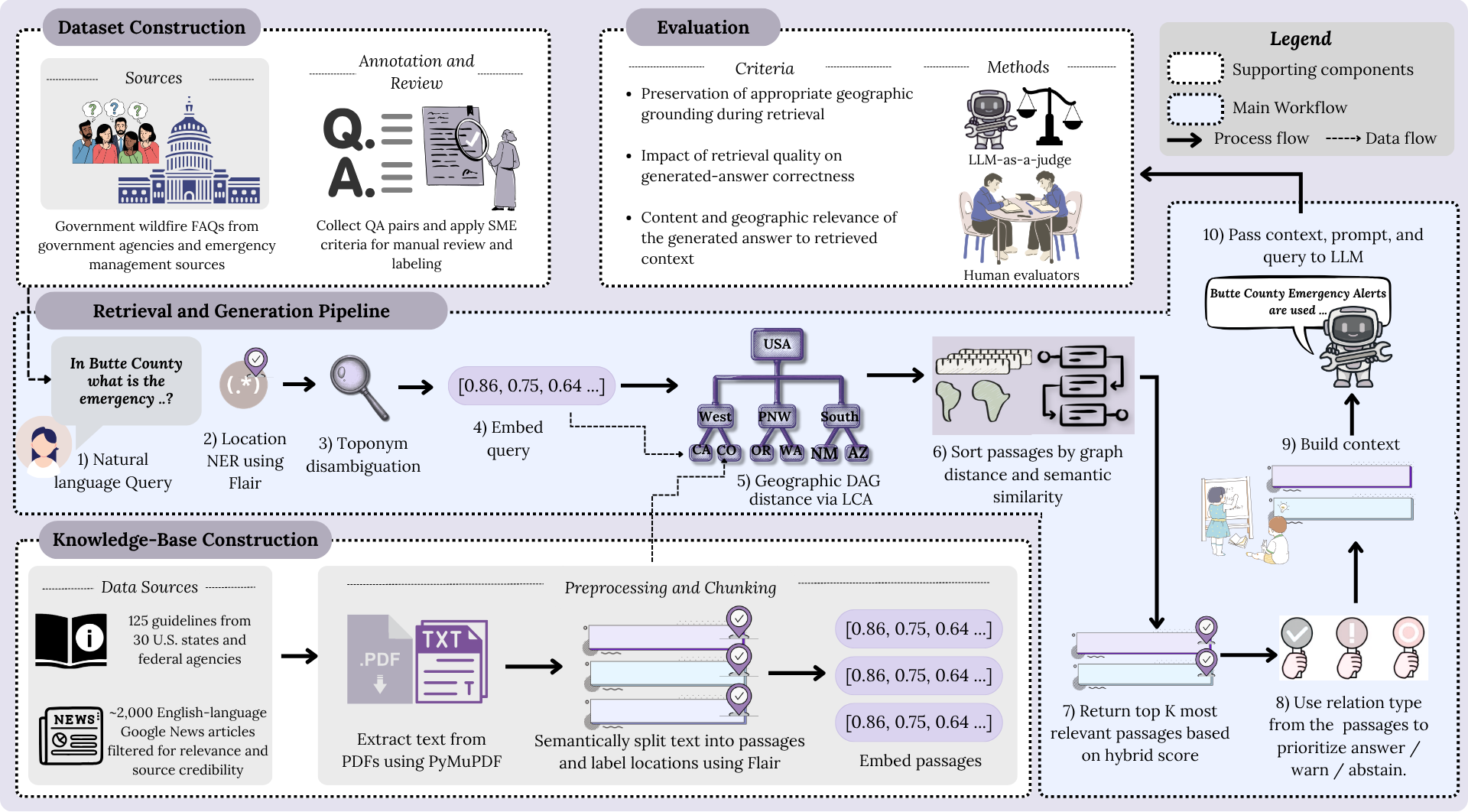}
    \caption{High-level framework of the proposed GeoRisk-RAG for hierarchy-aware QA. 
    }
    \label{fig:framework}
\end{figure*}

For semantic similarity search, passages with language that overlap with that of the query confound the applicability of generated responses by retrieving passages that maybe semantically relevant but geographically inapplicable. In contrast, spatial similarity search captures the geographic proximity between the query's intended location and that of the respective passage, overlooking the idea that responses valid in one region may not transfer to another one despite being proximate. For example, the city of Ashland, Oregon, strictly prohibits planting highly flammable trees and shrubs within 30 feet of any structure, while other cities like Bend, Oregon have no such rule \cite{cityofashlandAshlandWildfireMitigation2025}.

To address this limitation, we propose GeoRisk-RAG, illustrated in Figure~\ref{fig:framework} that improves context-based retrieval and supports selective answering under imperfect retrieval. Figure~\ref{fig:example} illustrates a need for such a system. While this may appear like an expected behavior, gracefully escalating to higher-level geographic context with a calibrated warning is preferable as state-level guidance is often applicable to its constituent counties, providing the user actionable information to further investigate \cite{xieWildfireGPTTailoredLarge2024}.

GeoRisk-RAG approaches RAG reliability by leveraging hierarchical granularity levels from an external knowledge graph, Wikidata \cite{vrandecicWikidataFreeCollaborative2014}, to measure the hierarchical distance between the target location of a query and the geographic scope of retrieved passages using a hierarchy-aware geographic knowledge graph represented as a directed acyclic graph (DAG). The extracted relationship is classified as an exact match, broader context, more-specific context, or different place-context. This classification guides selective answering in which GeoRisk-RAG directly answers when context exactly matches the query geography, warns when context is geographically related but broader or more specific than the requested scope, and abstains when retrieved context concerns a different place that only shares a broad ancestor with the query's intended location. This design allows semantic relevance to be distinguished from geographic applicability before answer generation. 

Unlike existing geospatial RAG systems that rely on flat dense retrievers or predefined spatial boundaries, GeoRisk-RAG explicitly models administrative hierarchies and geographic granularity extracted from an external knowledge graph. To the best of our knowledge, this is the first framework to leverage structural spatial topology for both context retrieval and selective answering in disaster-response QA for enhancing region-aware guidance.

The key contributions of this research are as follows:
\begin{enumerate}[nosep]
\item \textbf{GeoRisk-RAG Framework}: We propose a hierarchy- and granularity-aware RAG framework for natural-hazard related QA. By framing retrieval as a joint rank optimization task, GeoRisk-RAG considers both semantic relevance and geographic alignment to enhance location-based QA.
\item \textbf{Selective-Answering Strategy}: We develop a selective-answering strategy for geographic grounding. Rather than treating all semantically relevant passages as equally valid, GeoRisk-RAG restricts responses to exact geographic matches, issues warnings for broader or narrower granularities, and abstains when context refers to an entirely different jurisdiction.
\item \textbf{Novel Benchmark Dataset for Wildfire QA}: We construct a novel wildfire QA benchmark dataset comprising 449 diverse QA pairs that capture realistic public information needs, used to evaluate location-aware QA across varying geographic granularities.
\item \textbf{Comprehensive Evaluation}: We evaluate RAG reliability beyond semantic similarity by incorporating geographic validity metrics that assess correctness, answering behavior, context quality, and user preference.

\end{enumerate}

\section{Related Works}

\subsection{Hallucination Mitigation in RAG}
There have been extensive efforts to mitigate hallucinations, a common challenge in LLM answer generation, and improve reliability in such question-answering frameworks \cite{ajmalEvaluatingEffectivenessAdvanced2025}. Particularly, \cite{lewisRetrievalAugmentedGenerationKnowledgeIntensive2021}, which combines pre-trained parametric and non-parametric memory for answer generation, has been commonly used to ground generated answers in authoritative sources. AdaptiveCheck-GPT \cite{basuAdaptiveCheckGPTNovelSelfCorrecting2025} is a self-correcting framework that detects and mitigates hallucinations in real-time through dynamic prompt optimization. SelfCheck-GPT \cite{manakulSelfCheckGPTZeroResourceBlackBox2023} iteratively compares generated sentences against stochastically generated responses. A recent technique, Statement Accuracy Prediction, based on Language Model Activations (SAPLMA) \cite{azariaInternalStateLLM2023} builds classifiers, using an LLM’s hidden representations as inputs to predict the truthfulness of a sentence. Other human-in-the-loop-methods \cite{tarunHumanintheLoopSystemsAdaptive2025, chenImprovingRetrievalAugmentedGeneration2025} that leverage reinforcement learning have also been used. However, existing approaches remain largely general-purpose and do not explicitly consider domain-specific dependencies that determine whether retrieved context is applicable to a query.
\begin{figure}
    \centering
    \includegraphics[width=1.0\linewidth]{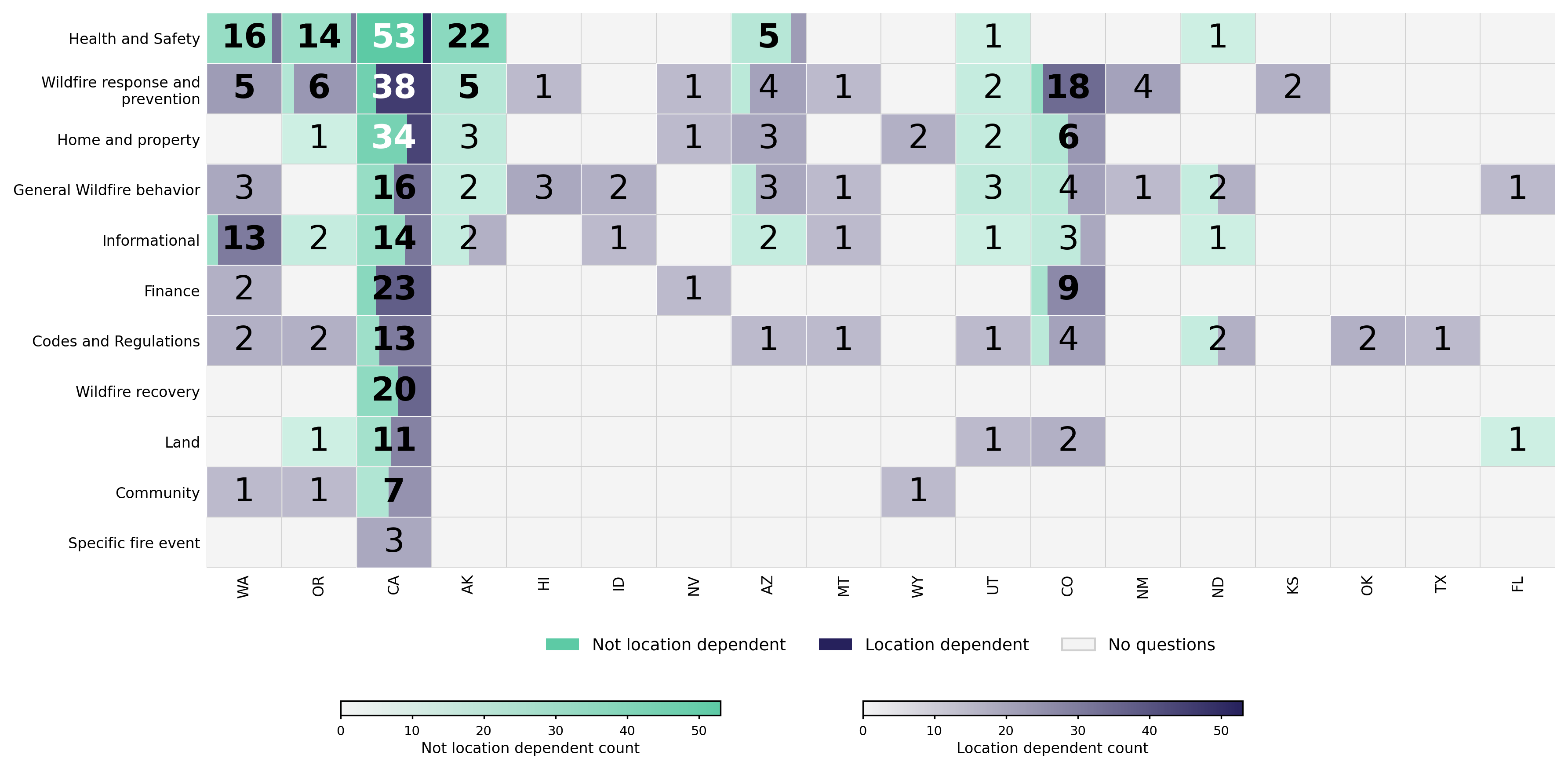}
    \caption{Distribution of the custom Wildfire QA benchmark dataset across states, question topics, and location dependency. Each row and column represents a question topic and a state respectively. 
    }
    \label{fig:datamap}
\end{figure}
\subsection{Graph-Augmented RAG Frameworks}
One particular approach to enhancing reliability in RAG frameworks using knowledge graphs is becoming increasingly prevalent in domain-specific QA. GraphRAG \cite{yuGraphRAGR1GraphRetrievalAugmented2026} and KG-RAG \cite{lindersKnowledgeGraphextendedRetrieval2025} augment traditional RAG by exploiting structural relationships inherent in document-bases for enriched semantic context for LLMs. Similarly, GeoRAG \cite{chenGeoRAGGeographicRetrieval2026, wangGeoRAGQuestionAnsweringApproach2025} has been widely adopted for urban spatial reasoning to model spatial proximity and evolutionary processes. Finally, HippoRAG \cite{gutierrezHippoRAGNeurobiologicallyInspired2025} treats the KG as an artificial hippocampus that stores connections between entities and text for efficient retrieval.
While these methods improve the reliability of RAG systems in geographic QA, they rely on spatial proximity rather than using geographical hierarchy, which would better distinguish proximate places having different regulations and guidance.

\subsection{Natural Hazard QA Systems}
In natural hazard domains specifically, RAG frameworks have become more ubiquitous for improving factual accuracy. By grounding responses in real data such as scientific literature, as WildfireGPT \cite{xieMARSHAMultiagentRAG2025} and DisasterResponseGPT \cite{goecksDisasterResponseGPTLargeLanguage2023} do, emergency guidance has become more reliable. SafeMate, for example, uses modular retrieval to enhance preparedness in active emergency scenarios~\cite{jiaoSafeMateModularRAGBased2025}, while retrieval-enhanced LLMs have been shown to streamline communication during emergency calls~\cite{otalLLMAssistedCrisisManagement2024}. Pairing RAG with domain-specific training, as in ClimateGPT~\cite{thulkeClimateGPTAISynthesizing2024} or with KGs~\cite{wangWildlifeLookupChatbotFacilitating2025, jiaoSafeMateModularRAGBased2025} enables complex reasoning over structured data. However, these tools are adapted for domain-experts' decision-making rather than the general public. As such, there remains a significant gap in public-centered QA benchmark datasets. 

While extensive research has focused on enhancing the reliability of LLM-based QA frameworks, limited work has explored applicability gaps that arise from geographic granularity dependencies, which are significant in disaster-related scenarios. We address this gap by introducing a geographic applicability and selective-answering behavior evaluated using a novel wildfire-related benchmark dataset. 

\begin{table}
\centering
\caption{Statistics on Custom Wildfire QA Dataset}
\label{tab:wildfiregpt_summary}
\renewcommand{\arraystretch}{0.9}
\footnotesize
\begin{tabular}{lrr}
\toprule
\textbf{Statistic} & \textbf{Count} & \textbf{Percent} \\
\midrule
Total question-answer pairs & 449 & -- \\
States covered & 19 & -- \\
Unique cities covered & 28 & -- \\
Avg tokens in Question   & 18 & -- \\
Avg tokens in Answer & 156 & -- \\
\midrule
\multicolumn{3}{l}{\textbf{Geographic granularity}} \\
\quad City-level & 208 & 46.3\% \\
\quad County-level & 59 & 13.1\% \\
\quad State-level & 177 & 39.4\% \\
\quad Other jurisdiction-level & 5 & 1.1\% \\
\midrule
\multicolumn{3}{l}{\textbf{Location dependency}} \\
\quad Location-specific  & 49.2\% \\
\quad General wildfire & 50.8\% \\
\bottomrule
\end{tabular}
\end{table}
\section{Methodology}
\subsection{Benchmark Dataset Creation}
While many geographic QA datasets are publicly available, to the best of our knowledge, there does not exist an open-source natural hazard QA dataset suitable for this experimental setup. For example DisasterVQA \cite{al-mohannadiDisasterVQAVisualQuestion2026}, MapQA \cite{liMapQAOpendomainGeospatial2025}, and Geomorphology QA \cite{chenGeoRAGGeographicRetrieval2026} are widely used in geospatial domains, but they lack labeled geographic granularity data, often focus on factual information (e.g., ``What is the capital of France?''), are primarily synthetically generated. and adapted for domain-experts. 
The benchmark dataset was designed to evaluate whether the retrieval system of a RAG pipeline can preserve geographically appropriate grounding during retrieval and subsequent answer generation.
Therefore, we construct a dataset sourced from government wildfire-related frequently asked questions (FAQs) consisting of 449 QA pairs. Each QA pair was manually annotated across four distinct dimensions, $A = \{a_g, a_t, a_l, a_p\}$, where:
\begin{itemize}[nosep]
    \item $a_g$ denotes geographic granularity, capturing administrative levels and spatial scopes across multiple regions to ensure geographic diversity. 
    \item $a_t$ denotes topical category, capturing diverse public information needs grouped by source FAQ headings (e.g., health, preparedness, evacuation, property, and financial assistance), as detailed in Table~\ref{tab:wildfire-categories} and Figure~\ref{fig:datamap}.
    \item $a_l$ denotes location dependence, a binary variable indicating whether a QA pair requires specific geographic context such that its factual correctness, applicability, or guidance changes under geographic substitution.
    \item $a_p$ denotes public information need. A QA pair is considered public-facing if it includes actionable (rather than pure factual recall) information needs intended for general residents, property owners, evacuees, or community members rather than for domain experts. 
\end{itemize}

The specific statistics of this dataset are presented in Table~\ref{tab:wildfiregpt_summary}.

\subsection{Constructing Knowledge Base}
\subsubsection{Knowledge-base data}
The retrieval document-base was constructed as a heterogeneous knowledge base combining documents from \begin{enumerate}[nosep]
    \item event-centric reporting, and
    \item stable institutionally-grounded and procedural guidance
\end{enumerate}
This design considers both temporal wildfire documentation and authoritatively-grounded rule-based guidance, as shown in open-domain retrieval problems, where heterogeneous corpora improve retrieval robustness by balancing recency-sensitive context with authoritative long-form grounding documents \cite{karpukhinDensePassageRetrieval2020, lewisRetrievalaugmentedGenerationKnowledgeintensive2020}.

For event-centric reporting, approximately 2,000 open-access English-language Google News articles (2023 -- 2025) were obtained using wildfire-related keywords (full list found in Appendix \ref{app:newsarticles}). The following measures were taken to ensure a high-quality document base:
\begin{itemize}[nosep]
    \item \textit{Credibility}: Irrelevant media (e.g., fictional and entertainment articles) were filtered out to restrict the corpus to credible institutional and journalistic outlets (e.g., .gov, .org, and established news agencies; see Appendix \ref{app:newsarticles}).
    \item \textit{Redundancy}: Duplicate articles were removed based on title and publication year to mitigate redundancy bias in retrieval.
\end{itemize}

To incorporate institutionally grounded guidance, 125 official local and federal agency guidelines containing preparedness materials, rules and regulations, and recovery resources from 30 U.S. states were collected. These PDF format files were processed using PyMuPDF (Fitz) library, selected due to its speed, flexibility, and accuracy \cite{adhikariComparativeStudyPDF, philippovichModernApproachesExtraction2025}. 
\subsubsection{Knowledge-base Chunking Strategy}
Given that natural hazard reporting and documentation are provided for general public consumption, these sources typically follow a structured, topic-driven format in which related information is organized under respective headings or sections. Notably, a semantic splitting strategy was employed.
We formulate document chunking as a semantic segmentation task over adjacent sentence embeddings.
Given a document consisting of ordered sentences $S = \{s_1, s_2, ..., s_n\}$, sentence embeddings were generated using the \textit{BAAI/bge-small-en-v1.5} embedding model\cite{xiaoCPackPackagedResources2023}.
For adjacent sentence pairs, cosine distance was computed as:
\[
d_i = 1 - \cos(s_i, s_{i+1})
\]
where:
\[
\cos(s_i,s_{i+1})
=
\frac{s_i \cdot s_{i+1}}
{\|s_i\|\|s_{i+1}\|}
\]
A semantic boundary was introduced when:
\[
d_i > \tau_d
\]
where $\tau_d$ denotes a document-specific adaptive threshold defined as the 95th percentile of the document-level cosine distance distribution. Given that adjacent sentences share high continuity, this boundary serves as a conservative outlier threshold, restricting segmentation to the top 5\% of major contextual shifts, preventing over-segmentation while capturing topic continuity (total: 12,975; 5,664 chunks from guidelines and 7,311 from news articles with 545 average tokens per node, and 304 median tokens per node). 
\subsection{Location Named Entity Recognition}
The automated location Named Entity Recognition (NER) pipeline exploits flair \cite{akbikFLAIREasytoUseFramework2019} due to its performance within this study as well as its reputability \cite{hiltmannNER4allContextAll2025, madureiraGeolocatingNewsExtreme2026}. Multiple NER models, namely Flair, dslim/bert-base-NER (BERT) \cite{tjongkimsangIntroductionCoNLL2003Shared2003}, tner/deberta-v3-large-ontonotes5 (DeBERTa) \cite{heDeBERTaV3ImprovingDeBERTa2023}, and spaCy \cite{spacy2020}, were evaluated for accuracy and speed, as shown in Table~\ref{tab:ner_performance}, on an out-of-sample labeled dataset (GeoVirus \cite{grittaWhichMelbourneAugmenting2018}, $n=229$) due to its analogous news reporting structure. The NER model selection process was formulated as a multi-objective trade-off between
\[
\mathcal{O} = \{\max(F_1), \min(\lambda)\}
\] where:
\begin{itemize}[nosep]
    \item $F_1$ denotes the entity extraction accuracy ($F_1$-score),
    \item $\lambda$ denotes the inference latency per text.
\end{itemize}
\begin{table}[htbp]
\centering
\footnotesize
\caption{Location NER Performance on Out-of-Sample GeoVirus Dataset (n=229 texts)}
\label{tab:ner_performance}
\begin{tabular}{lrrrr}
\toprule
\textbf{Metric} & \textbf{Flair} & \textbf{SpaCy} & \textbf{BERT} & \textbf{DeBERTa} \\
\midrule
Precision& 0.943 & 0.829 & 0.842 & 0.865 \\
Recall (Completeness)& 0.928 & 0.775 & 0.884 & 0.891 \\
F1-Score & 0.929 & 0.789 & 0.853 & 0.878 \\
Avg Latency (s) per Text& 0.592 & 0.062 & 0.128 & 0.744 \\
Total Execution Time (s)& 135.670 & 13.900 & 29.399 & 170.404 \\
\bottomrule
\end{tabular}
\end{table}

While spaCy exhibited fast performance (0.062s/text), it struggled to differentiate toponyms and ambiguous tokens (e.g., IN was extracted when used as a preposition). The BERT-based model, though slower than spaCy (0.128s/text), improved in accuracy by identifying additional geographical entities such as mountain ranges. Among the evaluated models, flair achieved the highest extraction quality ($F_1=0.9290$), while mitigating significant latency introduced by larger transformer models.

\subsection{Hierarchical Geographic Retrieval Framework}
\begin{table*}[h]
\centering
\small
\setlength{\tabcolsep}{4pt}
\caption{Example of GeoRisk-RAG behavior for an Oregon-state level wildfire query.}
\label{tab:hierarchy_relation_example}
\resizebox{\textwidth}{!}{%
\begin{tabular}{llcllc}
\hline
\multicolumn{6}{l}{\textbf{Example query:} ``How do wildfire risks vary in Oregon?''} \\
\multicolumn{6}{l}{\textbf{Query hierarchy:} \texttt{[Oregon, Pacific Northwest, United States]}} \\
\hline
\textbf{Passage type} & \textbf{Passage hierarchy} & \textbf{Dist.} & \textbf{Shared ancestor} & \textbf{Relation} & \textbf{Decision} \\
\hline
Exact Oregon passage & \texttt{[Oregon, Pacific Northwest, United States]} & 0 & Oregon & exact match & answer \\
Broader U.S. passage & \texttt{[United States]} & 2 & United States & passage broader than query & warn \\
Same-region passage & \texttt{[California, West, United States]} & 2 & West & different place, same ancestor & abstain \\
County-level passage & \texttt{[Multnomah County, Oregon, Pacific Northwest, United States]} & 1 & Oregon & passage more granular than query & warn \\
City-level passage & \texttt{[Portland, Multnomah County, Oregon, Pacific Northwest, United States]} & 2 & Oregon & passage more granular than query & warn \\
\hline
\end{tabular}%
}
\end{table*}

While recent works \cite{khodizadeh-nahariNovelSimilarityMeasure2021} have proposed a granular computing framework for measuring location similarity based on granularity levels, these approaches do not account for the hierarchical structure of geographic entities within a knowledge graph, nor have they applied to knowledge retrieval validation. We extend this foundation by grounding granularity levels within a Directed Acyclic Graph (DAG) derived from Wikidata \cite{vrandecicWikidataFreeCollaborative2014}, supporting a distance measure that captures both semantic similarity and spatial containment relationships, enhancing geographically precise responses.

We approach this problem of geographic validity by adopting a Granular Computing approach \cite{khodizadeh-nahariNovelSimilarityMeasure2021}, computing the similarity of two locations based on their granulation level using a graph structure. 
\subsubsection {Defining the DAG Model}
The Wikidata knowledge graph is used as the underlying geographic knowledge source, providing the graph structure over which granulation levels and location similarities are computed. 
The geographic knowledge space is represented as a DAG:
\[
G = (V,E)
\]
where $V$ denotes geographic entities, and $E$ denotes directed containment relationships.
Each entity $v \in V$ corresponds to a Wikidata QID.
Directed edges are constructed using Wikidata containment relations including:
\textbf{p131} for ``located in the administrative territorial entity'', and \textbf{p361} for ``part of''.

An example of how a query and passages are mapped to their respective hierarchies and relationships between the passage and question via the hierarchy is shown in Table~\ref{tab:hierarchy_relation_example}.
\subsubsection {Geographic Distance Measure}
We define the hierarchy $H(v)$ of any geographic entity $v$ as the set of all ancestors reachable via the transitive closure of edges: $H(v)={p \in V |(v,p) \in E^+}$ where $E^+$ is the transitive closure of $E$.
To compare a query location,$q \in V$, and a passage text location $t \in V$, the distance through their Lowest Common Ancestor (LCA) is calculated as:
$$\delta(q, t) = (i_q) + (i_t)$$
where:
\begin{itemize}[nosep]
\item The LCA is the first element in $H_q$ that is ordered from leaf to root and that also appears in $H_t$
\item $i_q$ and $i_t$ are the indices of the LCA in the set of $H_q$ and $H_t$ respectively.
\end{itemize}
In the case that either $H_q$ or $H_t$ is empty, or no shared ancestor exists, $\delta(q, t) = \infty$ and the passage is excluded from consideration. Structurally, this pipeline resolves 98.2\% of extracted geographic entities to precise Wikidata QIDs, while unresolvable entities gracefully default to the parent level within the DAG to prevent failures.
\subsubsection{Similarity Measure}
Candidate passages are evaluated and retrieved using a two-prong ordering scheme in which they are first sorted by minimizing geographic distance $\delta(q,t)$ and secondarily by semantic cosine similarity $\cos\theta$, as deterministic sorting criterion. We formalize this retrieval process as a rank optimization task. Defining the retrieved context window as a finite set of top-$k$ passages $\{t_1, t_2, \dots, t_k\} \in \text{candidate set P}$, the ordering objective using a  scoring function is as follows:
\[f:{P} \to \mathbb{R}^2\]
where:
\begin{itemize}[nosep]
    \item $f(t)=(-\delta(q, t), \, \cos\theta(q, t))$
    \item  $\cos \theta$ is the cosine of the angle between the query vector and passage vector: \[\cos \theta = \frac{\sum_{i=1}^{n} q_i t_i}{\sqrt{\sum_{i=1}^{n} q_i^2} \sqrt{\sum_{i=1}^{n} t_i^2}}\]
\end{itemize}

The objective is to find an optimal permutation $\sigma$ over the candidate set such that the output vectors are ordinally maximized: $\max_{\sigma} \sum_{i=1}^{k} f(\sigma(i))$ where $k=6$, selected based on the elbow point of the average semantic cosine similarity curve (see Appendix \ref{app:topk}). By framing retrieval as a permutation maximization, the framework ensures that both geographic constraints and semantic relevance are prioritized during context retrieval.

\begin{figure}
    \centering
    \includegraphics[width=1\linewidth]{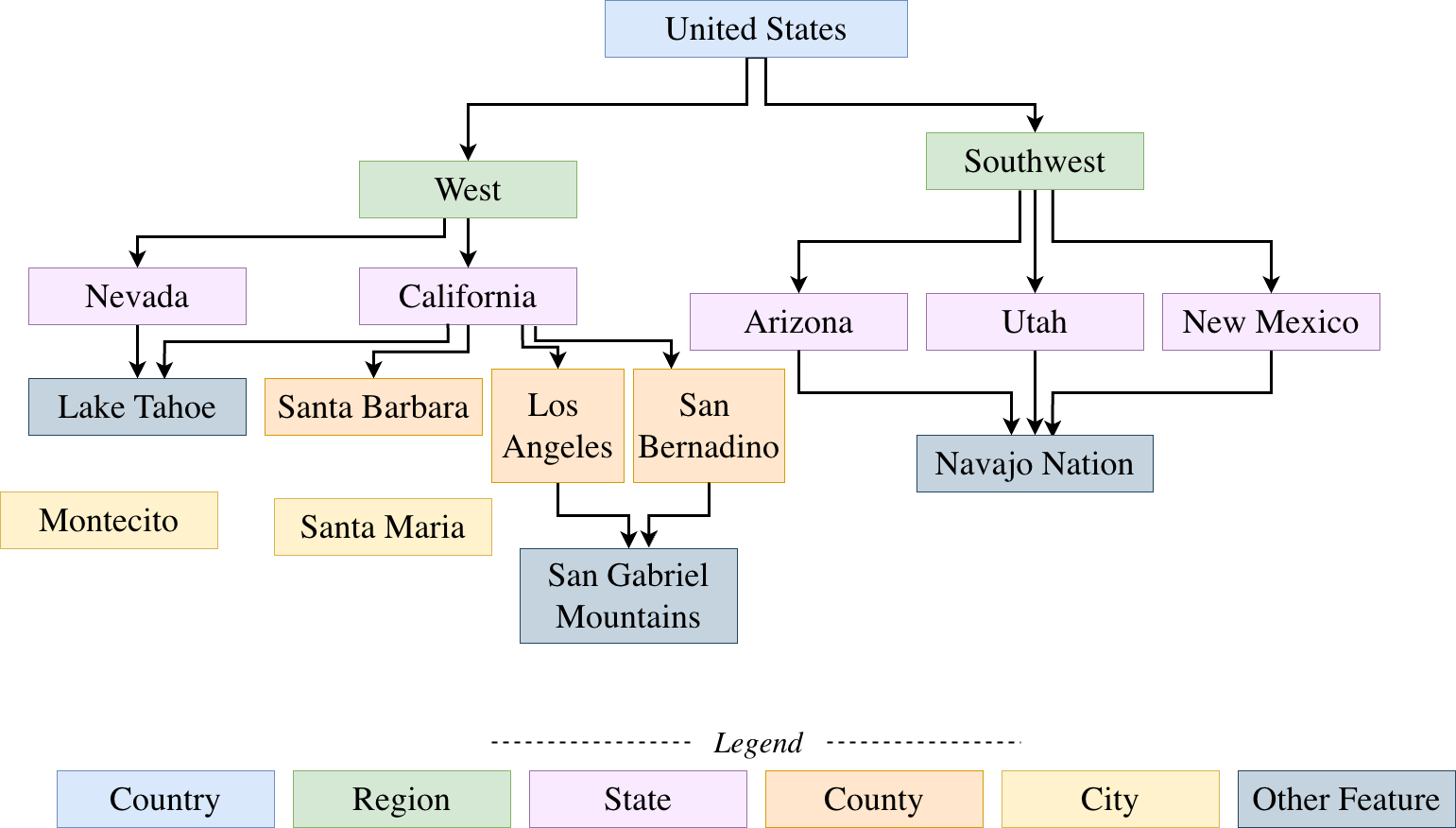}
    \caption{Example of geographical hierarchical DAG, illustrating the hierarchy structure and handling for ambiguous entities. Each color represents a granularity level.
    }
    \label{fig:ambiguous}
    \vspace{-1em}
\end{figure}

\subsection {Resolving Ambiguities}
The varying ways of expressing place names, nature of geographic references, and presence of non-standard geographic entities, necessitate disambiguation for improved retrieval. Therefore, several disambiguation concepts that drive how extracted locations are mapped to geographic entities in the graph are defined.
\subsubsection{Non-administrative Geographic Entities}
Non-administrative geographic entities (e.g., tribal lands and census-designated places) that do not always align with conventional administrative boundaries are mapped to their containing or nearest administrative equivalent (e.g., the corresponding county) while preserving original entity labels. This mapping provides clear hierarchical grounding, ensuring these regions enrich retrieval rather than being restricted.

\subsubsection{Landscape Features}
Hierarchies for natural features (e.g., rivers, mountain ranges, and parks) are constructed directly from Wikidata relations. For U.S.-based features lacking an explicit mapping in Wikidata, the hierarchy defaults directly to the country level as the baseline geographic context (see Figure~\ref{fig:ambiguous}).

\subsubsection{Aliases}
To improve coverage and recall, geographic aliases are resolved by traversing the ``alias'' metadata in Wikidata. This process maps variant identifiers (e.g., abbreviations, nicknames) directly to their canonical entities.
\subsubsection{Toponym Disambiguation}
Conversely, distinct spatial entities often share the same name (e.g., Springfield, IL vs. Springfield, MA). Therefore, the following prioritization scheme is applied to disambiguate such toponyms:
\begin{enumerate}[nosep]
    \item \textit{Explicit identifiers:} Candidates matching explicit state, country, or administrative metadata in the text are prioritized.
    \item \textit{Context:} The most frequently mentioned geographic entity contained within surrounding text is used, prioritizing candidates with the highest co-occurrence frequency. Ties defer to the third criterion.
    \item \textit{Population:} Remaining candidates are ranked by population (using Wikidata metadata), assuming more populous entities are more likely to be referenced implicitly.
    \item \textit{Sitelink counts:} As the final criterion, Wikidata sitelink counts serve as a proxy for prominence, prioritizing candidates with broader coverage across Wikimedia projects.
\end{enumerate}
While the current method used for entity disambiguation is exposed to the user, in practice, entities gleaned from each source of ambiguity above would be presented to the user as suggestions or elicit alternative clarification.

\section{Experimental Study}
\begin{table}[h]
\tiny
\centering
\caption{Summary of baseline methods used for comparison. Each method differs in its data structure, retrieval logic, and role in the study.}
\begin{tabularx}{\linewidth}{>{\raggedright\arraybackslash}X 
                              >{\raggedright\arraybackslash}X 
                              >{\raggedright\arraybackslash}X 
                              >{\raggedright\arraybackslash}X}
    \toprule
    \textbf{Method} & \textbf{Data Structure} & \textbf{Retrieval Logic} & \textbf{Purpose} \\
    \midrule
    Standard RAG & Text Embeddings & Vector cosine similarity only. & Standard, commonly used baseline. \\[4pt]
    Standard RAG + Cross Encoder & Reranked List & Semantic reranking of top 10 results. & Demonstrates that reranking alone is insufficient for geographic retrieval. \\[4pt]
    Granularity Match & Ordinal Scale & Matches based on geographic scope (city, county, state, region, country). & Highlights the need for relationships other than ancestor relationships. \\[4pt]
    Exact Keyword Match & String Literal & Text must match exactly (e.g., ``Texas'' == ``Texas''). & Shows insufficiency of literal string matching for varied geographic aliases. \\[4pt]
    GeoRisk-RAG & Directed Acyclic Graph & Knowledge graph logic using Wikidata for multi-parent resolution. & Resolves tribal lands, ambiguous places, and census-designated areas. \\
    \bottomrule
\end{tabularx}

\label{tab:baselines}
\end{table}

\begin{table*}[t!]
\centering
\caption{Reliability and selective-answering performance across geographically dependent and non-dependent wildfire QA tasks. GeoRisk-RAG achieves the strongest grounded decision behavior while substantially reducing false confidence.}
\label{tab:reliability}
\footnotesize
\resizebox{\textwidth}{!}{%
\begin{tabular}{lcccccc}
\toprule
\textbf{Method}  & \textbf{Good Decision Quality Rate } & \textbf{Faithfulness} & \textbf{Appropriate Warning Rate} & \textbf{Over-Abstention Rate} & \textbf{False Confidence}& \textbf{ Factual Correctness} \\
\midrule
\multicolumn{7}{l}{\textit{Location-Dependent Queries}} \\
\midrule
Standard RAG+Cross Encoder    & 0.751 (0.692--0.810) & 0.773 (0.693--0.852) &  0.001 (0.000--0.001) & 0.086 (0.054--0.127) &0.090 (0.054--0.131) &0.557 (0.493-0.620)\\
Standard RAG          & 0.819 (0.769--0.869) & 0.800 (0.714--0.876) & 0.009 (0.000--0.023) & 0.036 (0.014--0.063) & 0.095 (0.059-0.136)&0.679 (0.615-0.742) \\
Keyword Match      & 0.908 (0.896--0.939) & 0.909 (0.878--0.958) & 0.005 (0.000--0.014) & 0.045 (0.018--0.073) &0.023 (0.005--0.045)& 0.615 (0.552--0.683) \\
Granularity Match  & 0.824 (0.769--0.873) & 0.750 (0.666--0.845) &  0.001 (0.000--0.001) & 0.041 (0.018--0.068) &0.095 (0.059--0.136)& 0.561 (0.498--0.624)\\
\textbf{GeoRisk-RAG}& \textbf{0.946 (0.910--0.977)} &  \textbf{0.965 (0.912--1.000)}  & \textbf{0.195 (0.140--0.244)} & \textbf{0.030 (0.014--0.053) } & \textbf{0.009 (0.000--0.023)}&\textbf{0.761 (0.708--0.824) }\\

\midrule
\multicolumn{7}{l}{\textit{Non-Location-Dependent Queries}} \\
\midrule
Standard RAG+Cross Encoder  & 0.741 (0.689--0.798) & 0.814 (0.756--0.872) &  0.022 (0.004--0.044) & 0.079 (0.044--0.114) & 0.127 (0.083--0.175)&0.794 (0.750--0.846) \\
Standard RAG        & 0.711 (0.649--0.763) & 0.805 (0.738--0.866) &0.004 (0.000--0.013) & 0.061 (0.031--0.096) & 0.127 (0.083--0.171)	&0.772 (0.715--0.825) \\
Keyword Match    & 0.895 (0.855--0.930) & 0.951 (0.915--0.979) & 0.001 (0.000--0.001) & 0.039 (0.018--0.066) & 0.031 (0.013-0.053)&0.759 (0.702-0.820) \\
Granularity Match & 0.754 (0.697--0.811) & 0.787 (0.713--0.853) &  0.009 (0.000--0.022) & 0.039 (0.018--0.066) &0.127 (0.083-0.171)	& 0.728 (0.667--0.781) \\
\textbf{GeoRisk-RAG}& \textbf{0.974 (0.952--0.991)} & \textbf{0.963 (0.914--1.000)} &  \textbf{0.316 (0.254--0.382)} & \textbf{0.004 (0.000--0.013)} & \textbf{0.013 (0.000--0.031)}&\textbf{0.80 (0.737--0.838)} \\
\bottomrule
\end{tabular}%
}
\footnotesize{\textit{Values shown as \{mean (95\% CI).}\}}
\vspace{-1em}
\end{table*}
\subsection{Baselines and Models}
To evaluate the advantages of GeoRisk-RAG, four retrieval methods (summarized in Table~\ref{tab:baselines} and Table~\ref{tab:reliability}) spanning distinct assumptions across semantic similarity, lexical grounding, and geographic granularity are leveraged.
\begin{itemize}[nosep]
    \item Semantic Similarity \cite{lewisRetrievalAugmentedGenerationKnowledgeIntensive2021}: A standard dense RAG paradigm that retrieves the top-$K$ passages based on embedding similarity, serving as a baseline for location-agnostic semantic relevance.
    \item Standard RAG+Cross Encoder assisted Reranking \cite{jayavardhanaOptimizingRetrievalAugmentedGeneration2025}: Simultaneously processes a search query and a passage through a transformer model to predict a direct relevance score used for reranking. A lightweight \textit{ms-marco-TinyBERT-L4 \cite{petrovShallowCrossEncodersLowLatency2024}} model is used to evaluate if semantic reranking alone resolves geographic mismatch.
    \item Granularity only-based Retrieval: Restricts document retrieval strictly to the geographic granularity level of the query (e.g., state-level queries only retrieving state-level context) is used evaluate whether general scope matching is sufficient without explicit spatial alignment.
    \item Keyword matching: Matches explicit geographic entities expressed in the text to evaluate whether literal lexical grounding alone provides sufficient geographic validity.
\end{itemize}
For all of the experiments, two generative LLMs were used to generate responses (\textit{gpt-oss-120b} \cite{openaiGptoss120bAmpGptoss20b2025} and \textit{gemma-4-31B} \cite{teamGemmaOpenModels2024} See Appendix \ref{app:configsettings} for configuration settings). These models were chosen due to their advantages in different reasoning levels.

Furthermore, two transformer-based models (\textit{bge-small-en-v1.5} and \textit{bge-base-en-v1.5} \cite{bge_embedding}) were leveraged for embedding and retrieval tasks due to their open-source availability and computational efficiency. Both sizes were included to assess whether embedding capability influences retrieval performance under geographic constraints.
Together, these four combinations of embedding and generative models (\textit{bge-small-en-v1.5 + gpt-oss-120b, bge-small-en-v1.5 + gemma-4-31B, bge-base-en-v1.5 + gpt-oss-120b, and bge-base-en-v1.5 + gemma-4-31B}) form the set of experimental configurations evaluated in this work.

\begin{table}[H]
\centering
\footnotesize
\setlength{\tabcolsep}{6pt}
\caption{Overall stability analysis across RAG model configurations.}
\label{tab:model_robustness}
\begin{tabular}{lcc|cc}
\toprule
 & \multicolumn{2}{c|}{\textbf{bge-base-en-v1.5}} & \multicolumn{2}{c}{\textbf{bge-small-en-v1.5}} \\
\midrule
\multicolumn{5}{l}{\textit{Retrieved Passages (Semantic Retrieval Stability)}} \\
\midrule
Standard+Cross Encoder & 0.815 & & 0.816 & \\
Granularity Match      & 0.820 & & 0.829 & \\
Keyword Match          & 0.863 & & 0.872 & \\
Standard RAG           & 0.814 & & 0.821 & \\
\textbf{GeoRisk-RAG}   & \textbf{0.872} & & \textbf{0.873} & \\
\midrule
\multicolumn{5}{l}{\textit{Generated Answers (Semantic Generation Stability)}} \\
\midrule
& \makecell{\textbf{Gemma}\\\textbf{4-31B}} & \makecell{\textbf{GPT-OSS}\\\textbf{120b}} & \makecell{\textbf{Gemma}\\\textbf{4-31B}} & \makecell{\textbf{GPT-OSS}\\\textbf{120b}} \\
\cmidrule(lr){2-2} \cmidrule(lr){3-3} \cmidrule(lr){4-4} \cmidrule(lr){5-5}
Standard RAG+Cross Encoder & 0.735 & 0.712 & 0.738 & 0.729 \\
Granularity Match          & 0.753 & 0.706 & 0.752 & 0.730 \\
Keyword Match              & 0.804 & 0.777 & 0.806 & 0.773 \\
Standard RAG               & 0.733 & 0.692 & 0.717 & 0.707 \\
\textbf{GeoRisk-RAG}       & \textbf{0.838} & \textbf{0.821} & \textbf{0.842} & \textbf{0.814} \\
\bottomrule
\end{tabular}
\vspace{-1em}
\end{table}

\subsection{Model Configuration Stability Analysis}
To demonstrate that GeoRisk-RAG is not biased towards a particular model configuration, we conducted a stability analysis across the 4 configurations.

The robustness across configurations was measured using the mean cross-configuration pairwise cosine similarity ($\text{CC-CoSim}$) for a given configuration $c$ relative to all other combinations, leveraging an independent model (\textit{all-MiniLM-L6-v2} \cite{wangMiniLMDeepSelfAttention2020}) to mitigate bias:
\[\text{CC-CoSim}(c) = \frac{1}{|Q|} \sum_{q \in Q} \frac{1}{|C|-1} \sum_{c' \neq c} \cos\!\left(v(t_{q,c}),\, v(t_{q,c'})\right)\]
where $Q$ is the query set, $C$ is the configuration set, and $v(t_{q,c})$ denotes the dense embedding of text $t$ (i.e., passages or answers) under query $q$ and configuration $c$.

The consistent similarity scores across model configurations, as shown in Table~\ref{tab:model_robustness}, suggest that performance differences are driven by retrieval method rather than model choice. Hence, the subsequent experiments leverage \textit{bge-small-en-v1.5 + gpt-oss-120b} due to computational efficiency, with no meaningful drop in stability scores relative to bge-base-en-v1.5.

\subsection{Grounded Reliability and Correctness Behavior}

While LLMs are widely utilized for code generation, summarization, and refactoring, their application as a judge for verification and validation of natural language text is a developing technique \cite{sollenbergerLLM4VVExploringLLMasaJudge2024, liRAGZevalEnhancingRAG2025}. Standard text similarity metrics (e.g., cosine similarity) are strong at measuring lexical similarity, but often struggle at evaluating semantic validity of the response. Given both this technical limitation and the limited availability of domain experts for iterative manual evaluation, an LLM-as-judge framework was leveraged to assess reliability and quality of GeoRisk-RAG. We extend traditional RAG metrics to include selective-answering behavior under geographic uncertainty, where abstention and warning behavior may be preferable to confident but geographically invalid responses.

\subsubsection{Metrics}

To comprehensively evaluate GeoRisk-RAG, we adopt 6 metrics that balance retrieval and generation quality with geographic alignment.

\begin{itemize}[nosep]
    \item \textit{Good Decision Quality Rate (GDR)}: The proportion of cases in which the system makes an appropriate decision given the retrieved context. Appropriate decisions include providing a grounded answer when sufficient context is available, warning on geographic context limitations, or abstaining when the retrieved context is insufficient.

    \item \textit{Faithfulness}: The average proportion of generated answers strictly entailed by the retrieved context. 

    \item \textit{Appropriate Warning Rate}: The proportion of cases in which the system appropriately provides a warning when the retrieved context is insufficient, irrelevant, or geographically mismatched.

    \item \textit{Over-Abstention Rate}: The proportion of cases in which the system declines to answer despite the presence of sufficient, relevant context.

    \item  \textit{False confidence rate}: The proportion of cases in which the system answers confidently despite the context being geographically or semantically irrelevant (i.e., failing to warn or abstain). 
    
    \item \textit{Factual Correctness Rate}: The proportion of responses matching the ground-truth reference answer, capturing domain-specific accuracy and consistency.
    
\end{itemize}

\subsubsection{Evaluation Setup}
The \textit{Nemotron-3-Super} is used as the LLM judge due to its reported effectiveness in long-context reasoning \cite{bercovichLlamaNemotronEfficientReasoning2025}. For each QA pair, the judge is provided with five inputs: (i) the user query, (ii) the target geography extracted from the query, (iii) the ground-truth response, (iv) the retrieved context, and (v) the response generated by GeoRisk-RAG or a baseline model and prompted to assign labels corresponding to the 6 evaluation metrics.

These labels are aggregated across the evaluation set to assess GeoRisk-RAG's groundedness, correctness, and reliability under both geographically and non-geographically constrained queries. Additionally, for all these metrics, the 95\% confidence intervals is reported using bootstrap resampling over the evaluation queries. Specifically, the process of sampling queries with replacement while the respective metric is recomputed over the resampled set, is performed 1,000 times. 

The 5 and 95 percentiles of the resulting bootstrap distribution serve as the respective  lower and upper bounds of the confidence interval, quantifying the variability of the evaluation results relative to the sampled query set.

\subsection{Geographic Applicability Analysis}
To further explain how geographic context affects such decision choices, we focus on the relationship among the target location, the retrieved context, and the resulting answer behavior. It is crucial to distinguish these in natural hazard domain QA as they guide decision-making.

Therefore, we conduct two experiments to examine whether retrieval methods preserve geographic applicability, beyond topic relevance:
\begin{enumerate}[nosep]
    \item \textit{Decision Consistency:} How reliably do retrieval methods make appropriate selective-answering decisions across scenarios where answers are sensitive to geographic substitution?
    \item \textit{Error Sensitivity:} How does answer correctness vary across different types of geographic context failures, such as missing context, explicit geographic mismatches, and misaligned spatial granularities (broader or narrower scopes)?
\end{enumerate}

\subsubsection{Geographic errors}

To assess geographic vulnerabilities in context retrieval, each response is evaluated using LLM-as-Judge to identify the geographic error type based on the relationship between the intended geography expressed in the query and that of the retrieved context. The following categories are leveraged:

\begin{itemize}
    \item \textit{Exact Geo Match}: the retrieved context is geographically appropriate for the query. This includes exact geographic matches or context that is applicable for the target location.

    \item \textit{No Geo context}: the retrieved context does not contain sufficient geographic context to support an answer for the target location.

    \item \textit{Geo Mismatch}: the retrieved context refers to a different place that has weak geographically applicability to the target location despite semantically relevant. This accounts for sibling and cousin relationships in the hierarchy (i.e., if two locations share an LCA but neither is an ancestor of the other, the evidence is accounted within this category). 

    \item \textit{Coarse Geo}: the retrieved context applies to more coarse geographic granularity than the target location. Selective answering should typically trigger a warning rather than a completely confident answer.

    \item \textit{Fine Geo}: the retrieved context applies to a more granular geographic scope than the target location. Such should also typically require a warning or abstention depending on tenability of the context.
\end{itemize}
\begin{table}[t!]
\centering
\footnotesize
\caption{Response correctness by geographic error type across retrieval methods.}
\label{tab:geo-error-correctness}
\setlength{\tabcolsep}{3pt}
\begin{tabular}{lccccc}
\toprule
\makecell{\textbf{Geographic-}\\\textbf{Error Type}} 
& \makecell{\textbf{Standard+}\\\textbf{Cross}\\\textbf{-Encoder}} 
& \makecell{\textbf{Keyword}\\\textbf{Match}} 
& \makecell{\textbf{Granularity}\\\textbf{Match}} 
& \makecell{\textbf{Standard}\\\textbf{RAG}} 
& \makecell{\textbf{GeoRisk-}\\\textbf{RAG}} \\
\midrule
Exact Geo Match        & 0.718 & 0.745 & 0.747 & 0.683 & 0.811 \\
Geo Mismatch & 0.732 & 0.700 & 1.000 & 0.686 & 0.685 \\
No Geo context     & 0.350 & 0.420 & 0.419 & 0.472 & 0.357 \\
Coarse Geo           & 0.812 & 0.889 & 0.867 & 0.889 & 1.000 \\
Fine Geo        & 0.500 & 1.000 & 1.000 & 0.723   & 1.000 \\
\bottomrule
\end{tabular}
\vspace{-1em}
\end{table}

To evaluate how the correctness of generated responses change under different types of geographic evidence failures, we categorize each retrieved context according to its geographic relationship to the target location and obtain the proportion of responses judged correct within each of the 5 geographic error categories as shown in Table~\ref{tab:geo-error-correctness}.

\subsubsection{Location-dependent answer validity}
To better evaluate the conditions under which GeoRisk-RAG is effective, as it is important that the same type of question may require different evidence depending on the precise requested location,we evaluate how answer behavior changes as geographic area is substituted.
This is done by semantically grouping same wildfire questions (e.g., ``What is the alert system for wildfires in Butte County, CA'' and `What is the system used to alert the public during wildfires in Los Angeles, CA''. The GDR is computed for each baseline method within for each question type. This evaluates whether a method consistently answers, warns, or abstains appropriately across geographically sensitive question types, rather than succeeding only on isolated queries.

\subsection{Qualitative Human Evaluation}
GeoRisk-RAG was further evaluated by two annotators. For each of 50 randomly selected location-dependent questions, annotators were presented the original question and the unique responses generated by the 5 evaluation configurations (four baselines and GeoRisk-RAG). Method identities were concealed, and response order was randomized to mitigate bias. Annotators were asked to select one response that they preferred most based on its overall usefulness, clarity of guidance, and transparency of geographic risk (see Appendix~\ref{app:annotatorprompt}). Across all scenarios (excluding instances where all methods abstained), both annotators consistently preferred the response behavior of GeoRisk-RAG, achieving high inter-rater consensus (Cohen's $\kappa = 0.76$), indicating meaningful agreement. Annotator 1 selected it in 70.0\% of cases over the strongest baseline, while annotator 2 selected it in 68.75\% of cases. This suggests that considering structural geographic alignment improves the reliability of generated answers in natural hazard domains.

\section{Results}
As detailed in Table~\ref{tab:reliability}, GeoRisk-RAG achieves the highest overall reliability with a GDR of $0.946$ on location-dependent queries and $0.974$ on non-location-dependent queries, demonstrating the value of geographic reasoning during retrieval. Specifically, the improvement is most evident for location-dependent questions, where geographic applicability is significant. GeoRisk-RAG reduces the false confidence rate from $0.090$ to $0.009$ compared to standard semantic similarity and reranking baselines. Even on general wildfire queries (non-location dependent), GeoRisk-RAG remains competitive ($0.963$ GDR), highlighting that geographic constraints do not degrade performance on location-agnostic queries. Furthermore, the framework provides appropriate warnings, indicating that it distinguishes localized evidence from geographically coarse, fine, or irrelevant evidence rather than simply over-abstaining, while maintaining strong factual correctness ($0.761$).
\begin{figure}[H]
    \centering
    \includegraphics[width=\linewidth]{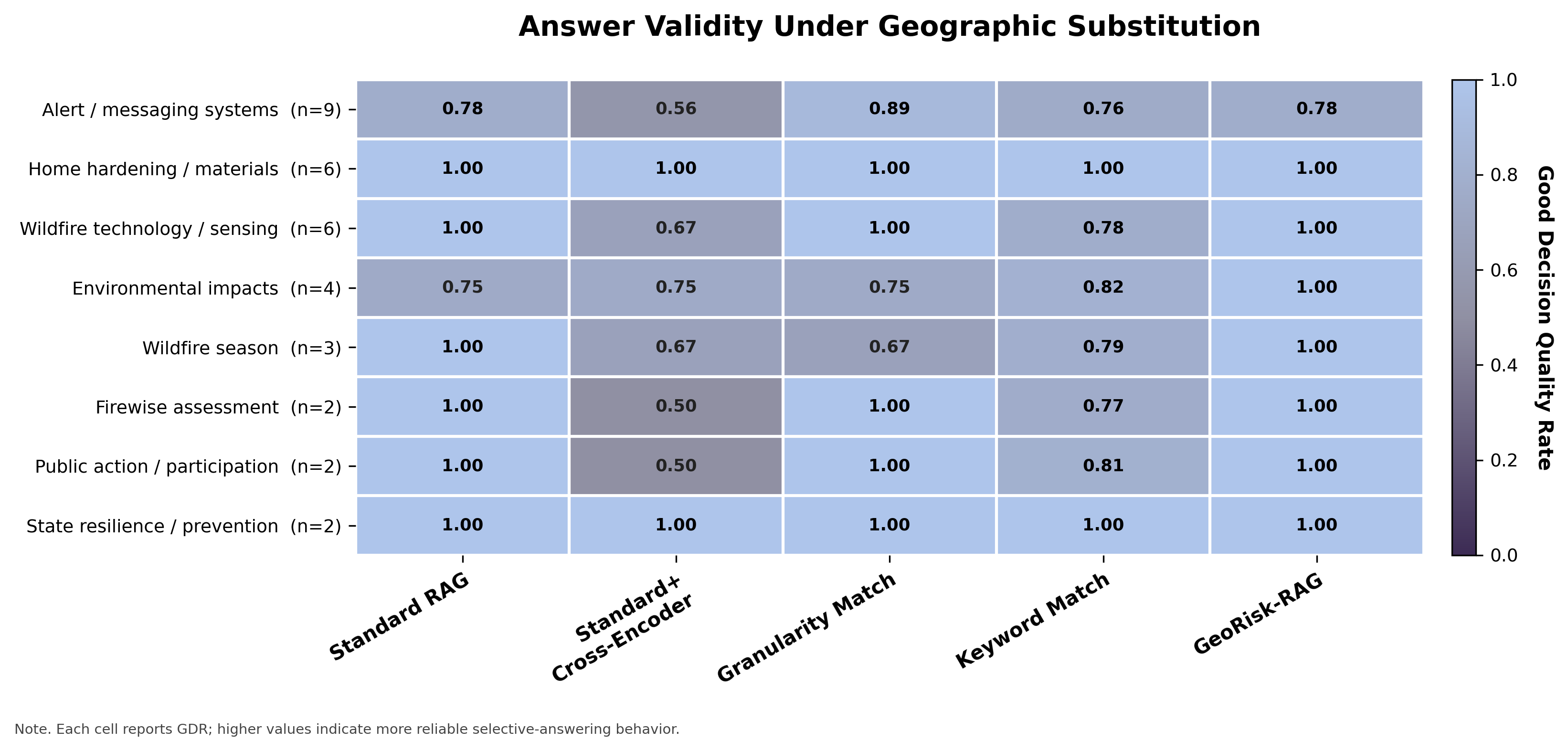}
    \caption{GDR across same questions with different locations to evaluate answering behavior. Higher values indicate more reliable selective-answering behavior.}
    \label{fig:geoheatmap}
    \vspace{-1em}
\end{figure}
When the same type of query is prompted across different locations, GeoRisk-RAG maintains high GDR across seven out of eight question types (Figure~\ref{fig:geoheatmap}). This is evident in questions inquiring about the timeframes of wildfire seasons, which require retrieving context at specific granularities, as answers vary even within a single state.

Table~\ref{tab:geo-error-correctness} illustrates how response correctness varies significantly with the type of geographic context available. Cases lacking geographic evidence have the lowest correctness ($0.404$ averaged across all five configurations), indicating appropriate abstention and highlighting the importance of geographic grounding. While broader or narrower evidence, which may not align with the target geography, can still support correct responses, baseline architectures like Cross-Encoders require selective-answering mechanisms to mitigate this risk.

\section{Case Study}
\begin{figure}
\centering
\includegraphics[width=1\linewidth]{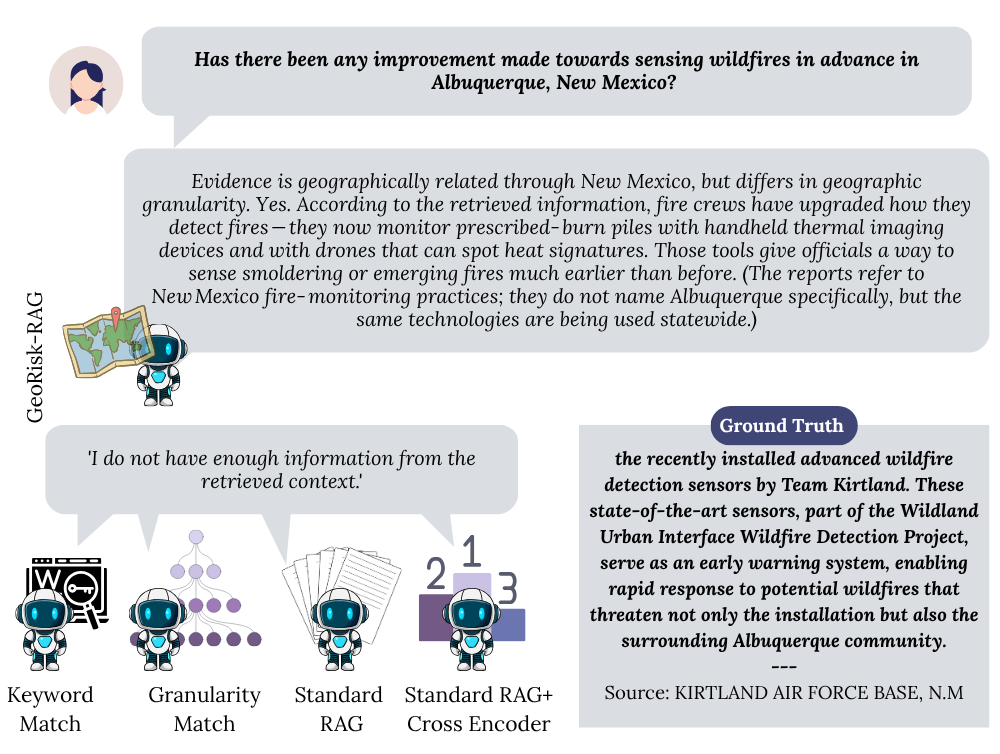}
\caption{A location-specific query about Albuquerque, New Mexico highlights GeoRisk-RAG's reasoning: rather than abstaining as all baselines do, the system retrieves state-level context and responds (which correctly aligns with presented ground truth answer) with a warning.}
\label{fig:casestudy}
\vspace{-1.5em}
\end{figure}

A case study was conducted to examine how GeoRisk-RAG enhances location-specific question-answering, as illustrated in Figure~\ref{fig:casestudy} where a user asks ``Has there been any improvement made towards sensing wildfires in advance in Albuquerque, New Mexico?''.

While there are no passages in the knowledge base relevant to wildfire sensing technologies in Albuquerque, New Mexico, GeoRisk-RAG gracefully escalates to the state level (New Mexico) and explicitly warns the user that the retrieved information applies at the state level, before providing a contextually relevant answer that aligns with the ground truth and was manually verified against the retrieved sources.

\section{Discussion and Conclusion}
In this work, we propose GeoRisk-RAG, a hierarchy-aware framework designed to enhance location-aware question-answering in natural hazard scenarios. By distinguishing semantic similarity from explicit geographic applicability via a Wikidata-derived DAG, this framework introduces an explainable selective-answering behavior for increased transparency to enhance decision-making. 

Experimental results on our novel 449-sample wildfire QA dataset demonstrate that GeoRisk-RAG significantly minimizes false confidence on location-specific queries relative to a standard baselines and alternate reranking strategies, while maintaining high factual correctness. Additionally, during blind human-evaluations, GeoRisk-RAG responses were selected as more preferable compared to other baseline responses.

While the framework has demonstrated effectiveness in this domain, this work posits avenues for future research. First, given that our benchmark dataset is modest in size in its current state and U.S.-based, expanding this to encompass other countries, languages, and natural hazards is critical for more equitable public-use deployments. Furthermore, the framework relies heavily on the coverage and correctness in external knowledge graphs like Wikidata. Future work will include methods of resolving absences in Wikidata, domain-expert feedback, and human-in-the-loop involvement for clarification during disambiguation. 

\href{https://github.com/meenuravi18/wildfire_research_project} {The code base, benchmark dataset, and experiments can be found at this GitHub repository.}

\section{Acknowledgments}
The author's affiliation with The MITRE Corporation is provided for identification purposes only, and is not intended to convey or imply MITRE's concurrence with, or support for, the positions, opinions or viewpoints expressed by the author.

\section{GenAI Usage Disclosure}
LLMs were used as part of the experimental methodology. Separately, ChatGPT-5 was used only to review the manuscript for grammatical and typographical errors and was not used for technical content generation or text generation.

\bibliographystyle{ACM-Reference-Format}
\bibliography{references}
\appendix

\section{Wildfire QA Dataset}
\begin{table}[H]
\centering
\tiny
\renewcommand{\arraystretch}{1.0}
\caption{Wildfire Question Categories}
\begin{tabularx}{\columnwidth}{>{\raggedright\arraybackslash\linespread{0.8}\selectfont}m{0.25\columnwidth} >{\raggedright\arraybackslash\linespread{0.8}\selectfont}m{0.5\columnwidth} >{\centering\arraybackslash}m{0.04\columnwidth}}
\hline
 \textbf{Topic} & \textbf{Description} & \textbf{Count}  \\
\hline
General Wildfire Behavior & Causes, spread, fire science, smoke behavior, seasons, risk factors. & 42\\
\hline
 Home and Property & Protecting homes, defensible space, structures, insurance for homes, property damage. & 52\\
\hline
 Codes and Regulations & Laws, building codes, permit rules, compliance requirements, official regulations. & 29\\
\hline
 Specific Fire Event & Questions about a named wildfire, a particular incident, location-specific active or past fire. & 3\\
\hline
 Community & Questions about how the community can help, community-level impacts.& 10 \\
\hline
 Land & Forests, vegetation, soil, erosion, watersheds, land management, acreage, habitat. & 16 \\
\hline
 Finance & Grants, loans, compensation, taxes, financial aid, business loss, economic cost. & 35\\
\hline
Wildfire Response and Prevention & Evacuation, preparedness, mitigation, fuel reduction, emergency response. & 89\\
\hline
 Wildfire Recovery & Debris removal, rebuilding, restoration, post-fire recovery steps, returning after fire.& 20 \\
\hline
 Health and Safety & Smoke exposure, breathing, injury, health risks, safety precautions for people. &112 \\
\hline
 Informational & General factual or definitional questions that do not fit the other categories strongly. & 41\\
\hline
\end{tabularx}
\label{tab:wildfire-categories}
\vspace{-1em}
\end{table}

\section {News Article Retrieval} \label{app:newsarticles}
The news domains reported in Table~\ref{tab:reputable_domains} include sources with established credibility ratings and minimal bias \cite{yangAccuracyPoliticalBias2025, kongAnalyzingGeographicBias2026}. Additionally, the following terms were used to query Google News and retrieve relevant news articles: ``wildfire'', ``forest fire'', ``bushfire'', ``brushfire'', ``fire season'', ``wildland fire'', ``wildland-Urban Interface''.
\begin{table}[H]
\centering
\tiny
\caption{Reputable news domains used for filtering Google News articles.}
\label{tab:reputable_domains}
\begin{tabularx}{\columnwidth}{p{0.30\columnwidth} X}
\hline
\textbf{Category} & \textbf{Domains} \\
\hline
Press Association & apnews.com, reuters.com, afp.com, upi.com \\
\hline
National U.S. & nytimes.com, washingtonpost.com, wsj.com, usatoday.com, npr.org \\
\hline
Regional & chicagotribune.com, latimes.com, bostonglobe.com, dallasnews.com, sfchronicle.com, denverpost.com, seattletimes.com, startribune.com, tampabay.com, miamiherald.com, inquirer.com, oregonlive.com \\
\hline
International & theguardian.com, ft.com, economist.com, abc.net.au, cbc.ca \\
\hline
Investigative & propublica.org, theatlantic.com \\
\hline
Fire / Disaster & wildfiretoday.com \\
\hline
Government & *.gov, *.org \\
\hline
\end{tabularx}
\end{table}
\section{Annotator Evaluation Prompt}\label{app:annotatorprompt}
The following prompt was provided to both annotators along with the 50 randomly selected questions:
\textit{For each question, choose the answer you would most prefer as a user. A strong answer should directly answer the question, provide useful and accurate guidance, and include an appropriate warning or abstain when the answer cannot be determined. If two answers seem equally good, choose the one that you would prefer if you were asking the question for the location.}
\section{Top-K Retrieval}\label{app:topk}
\begin{figure}[H]
    \centering
    \includegraphics[width=\linewidth]{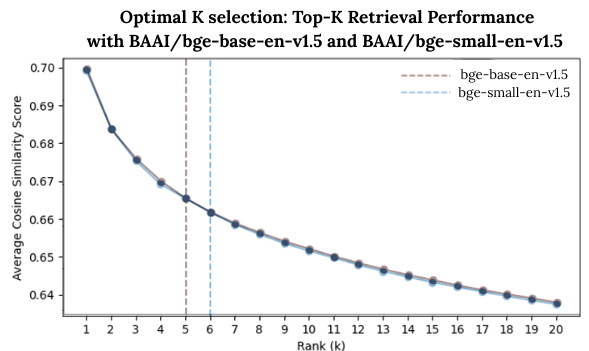}
    \caption{Optimal top-k selection using elbow-point method performed on wildfire benchmark dataset. 
    }
    \vspace{-1em}
\end{figure}

\section{Prompting Setup and LLM Parameters}\label{app:configsettings}
\begin{table}[h]
\centering
\tiny
\caption{LLM Models Configuration Settings}
\begin{tabular}{l|c|c|c}
\hline
\textbf{Model Name} & 
\makecell{\textbf{gpt-oss-120b}} & 
\makecell{\textbf{gemma-4-31B}} & 
\makecell{\textbf{nemotron-3} \\ \textbf{super}} \\
\hline

Params            & 120B              & 31B               & 120B \\

Context Length    & 131,072           & 262,144           & 1,048,576 \\

Weight Format     & MXFP4             & BF16              & FP8 \\

GPU Resources     & 4x A100 40GB      & 2x H200 141GB     & 4x H200 141GB \\

Temperature       & 0.1               & 0.1               & 0 \\

Max Tokens        & 800               & 800               & 200 \\
Top P             & 1.0               & 1.0               & 1.0 \\

Top K             & -1                & -1                & -1 \\

Repetition Penalty & 1.0              & 1.0               & 1.0 \\

Presence Penalty  & 0.0               & 0.0               & 0.0 \\

Frequency Penalty & 0.0               & 0.0               & 0.0 \\
Inference Framework & vLLM & vLLM & vLLM\\
Purpose        & \makecell{Benchmark} & \makecell{Benchmark }               & LLM-as-a-Judge \\
\hline
\end{tabular}%

\label{tab:model-comparison}
\end{table}

\begin{figure}[H]
    \centering
    \includegraphics[width=\linewidth]{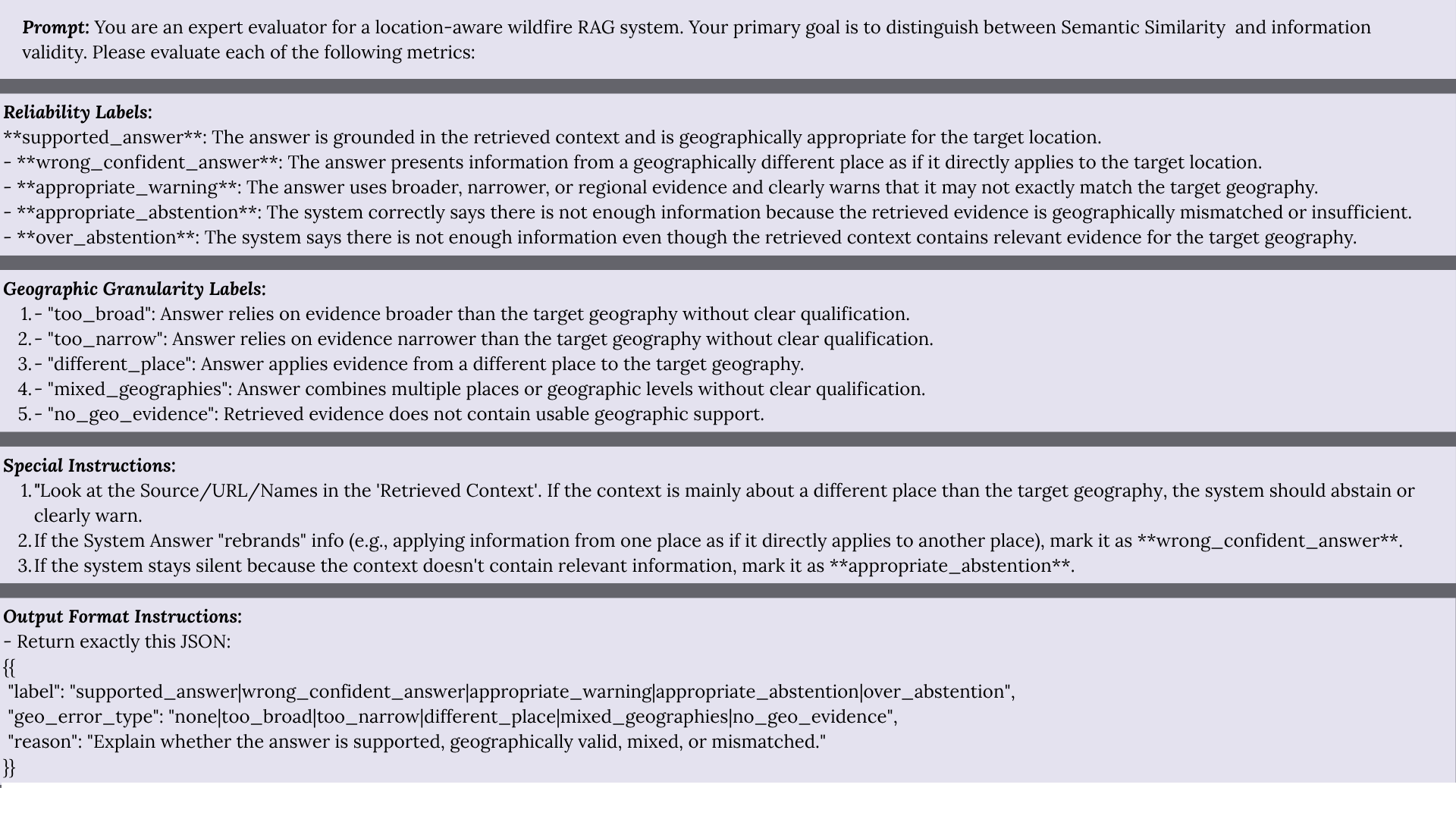}
    \caption{Prompt provided to nemotron-3-super LLM, urging it to act as an expert evaluator for a location-aware wildfire RAG system.
    }
    \label{fig:llmprompt}
\end{figure}

\end{document}